\documentclass[accepted]{uai2025} % after acceptance, for a revised version; 
\usepackage[american]{babel}
\usepackage{amsfonts}
\usepackage{amssymb}
\usepackage{natbib} % has a nice set of citation styles and commands
\usepackage{mathtools} % amsmath with fixes and additions
\usepackage{subfig}
\usepackage{booktabs} % commands to create good-looking tables
\usepackage{tikz} % nice language for creating drawings and diagrams

\usepackage{xurl}

\newtheorem{definition}{Definition}
\newtheorem{assumption}{Assumption}

\newtheorem{theorem}{Theorem}

\DeclareMathOperator*{\argmax}{arg\,max} % Jan Hlavacek

\title{Instance-Wise Monotonic Calibration by Constrained Transformation}

\author[1]{\href{mailto:<yunrui.zhang@unsw.edu.au>?Subject=Your UAI 2025 paper}{Yunrui Zhang}{\textsuperscript{*}}} 
\author[1]{\href{mailto:<g.batista@unsw.edu.au>?Subject=Your UAI 2025 paper}{Gustavo Batista}}
\author[1]{\href{mailto:<salil.kanhere@unsw.edu.au>?Subject=Your UAI 2025 paper}{Salil S. Kanhere}}
\affil[1]{%
    School of Computer Science and Engineering\\
    University of New South Wales\\
    Australia
}

\begin{document}
\maketitle
\renewcommand{\thefootnote}{\fnsymbol{footnote}}
\footnotetext[1]{Corresponding author: yunrui.zhang@unsw.edu.au}
\renewcommand{\thefootnote}{\arabic{footnote}}

\begin{abstract}
Deep neural networks often produce miscalibrated probability estimates, leading to overconfident predictions. A common approach for calibration is fitting a post-hoc calibration map on unseen validation data that transforms predicted probabilities. A key desirable property of the calibration map is instance-wise monotonicity (i.e., preserving the ranking of probability outputs). However, most existing post-hoc calibration methods do not guarantee monotonicity. Previous monotonic approaches either use an under-parameterized calibration map with limited expressive ability or rely on black-box neural networks, which lack interpretability and robustness. In this paper, we propose a family of novel monotonic post-hoc calibration methods, which employs a constrained calibration map parameterized linearly with respect to the number of classes. Our proposed approach ensures expressiveness, robustness, and interpretability while preserving the relative ordering of the probability output by formulating the proposed calibration map as a constrained optimization problem. Our proposed methods achieve state-of-the-art performance across datasets with different deep neural network models, outperforming existing calibration methods while being data and computation-efficient. Our code is available at \url{https://github.com/YunruiZhang/Calibration-by-Constrained-Transformation}
\end{abstract}

\section{Introduction}\label{sec:intro}

Deep learning models have achieved state-of-the-art accuracy across various tasks and applications. These models typically employ a softmax layer at the output, where the resulting outputs are interpreted as a probability distribution. However, despite their high accuracy, deep neural networks often suffer from miscalibration due to overfitting, producing inaccurate probability estimates. As demonstrated in prior work, these softmax-derived probabilities frequently fail to reflect true uncertainty, leading to systematically overconfident predictions~\citep{guo_calibration_2017}.

Accurate probability estimates are critical for safety-sensitive applications where model confidence serves as a failsafe mechanism or informs human decision-making. For example, in medical diagnosis, miscalibrated predictions could lead to unwarranted trust in automated results, risking misdiagnosis~\citep{van2019calibration}; in autonomous driving, overconfidence in object detection systems might compromise collision avoidance \citep{bieshaar2018cooperative}; and in financial decision-making, poorly calibrated uncertainty estimates could propagate costly errors~\citep{tang2014effects}, and many other application domains where an accurate estimate of the uncertainty is crucial for the task \citep{zhang2025revisit, tan2025pathsovergraph}. Beyond these domains, downstream tasks such as label shift estimation~\citep{alexandari_maximum_2020, saerens_adjusting_2002, zhang2025label} and out-of-distribution detection~\citep{lee2018training, hendrycks2017a} depend heavily on accurate probability estimates.

A common approach to improving the calibration of deep learning models is post-hoc calibration~\citep{guo_calibration_2017}. This approach assumes access to a validation set, referred to as the calibration set, consisting of data not seen during the training of a given base model. A parameterized calibration model is then trained on the outputs of the base model using this calibration set. The goal is to learn a calibration map that transforms the model’s probabilistic outputs to improve their alignment with true confidence levels.

Post-hoc calibration operates by adjusting a model’s predicted probabilities through a learned calibration map that modifies confidence scores based on the base model's predictions on the unseen calibration set. Specifically, when the model’s prediction is incorrect in the calibration set, the calibration map learns to reduce confidence in the misclassified class, whereas, for correct predictions, it increases confidence. This adjustment is achieved by optimizing the calibration map to minimize the loss between the adjusted probabilities and the ground-truth calibration set labels. Empirically, a well-learned calibration map can generalize effectively from the calibration to the test set.

Current post-hoc calibration methods can be broadly categorized into instance-wise monotonic and instance-wise non-monotonic methods, which we refer to as monotonic and non-monotonic. Monotonic methods preserve the ranking of predicted probabilities for each instance, ensuring that for each sample, the transformed probabilities maintain the same order as those produced by the original base model. In contrast, non-monotonic methods do not enforce this constraint.

\begin{figure}[htb]
    \centering
    \subfloat[Accuracy decrease]{\includegraphics[width= .5\columnwidth]{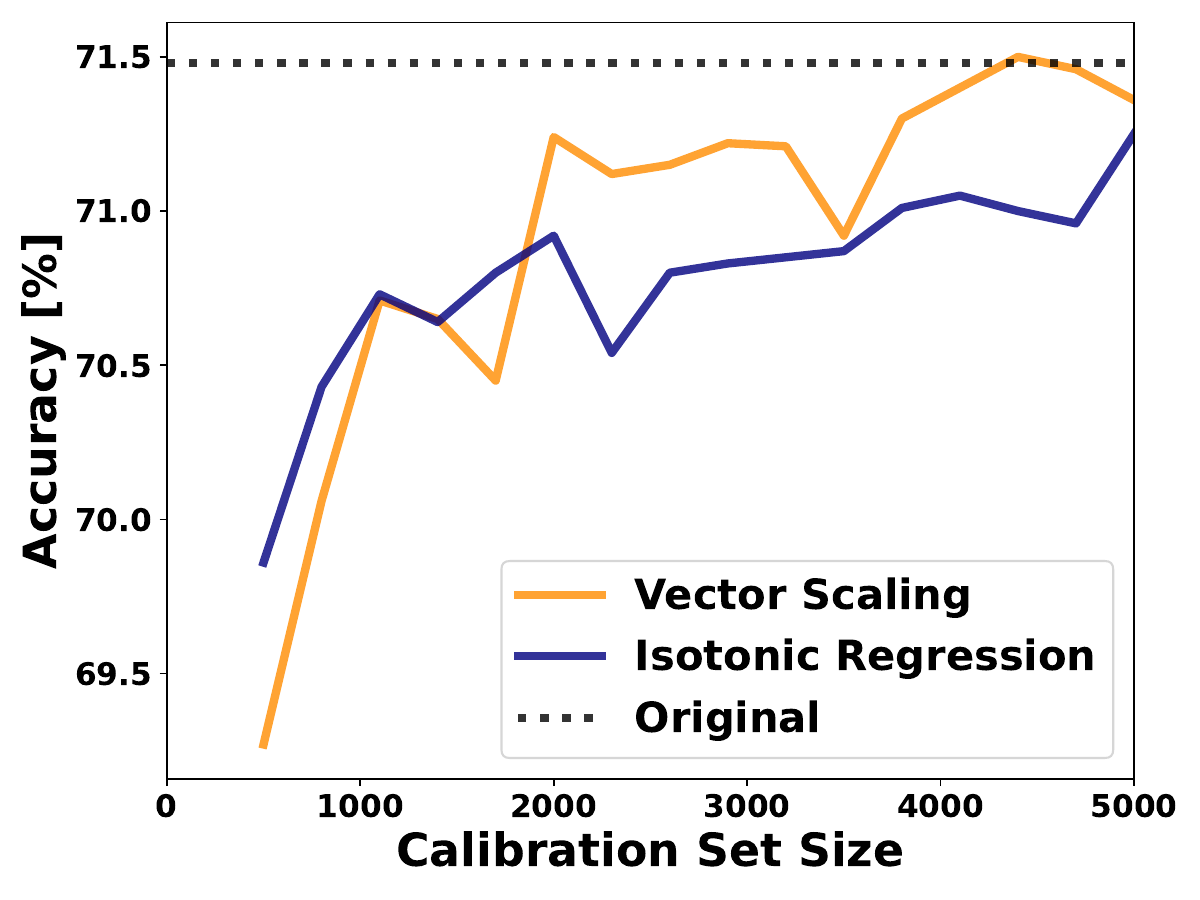}
    \label{fig:acc_dec}}
    \subfloat[Impact on uncertain samples]{\includegraphics[width= .5\columnwidth]{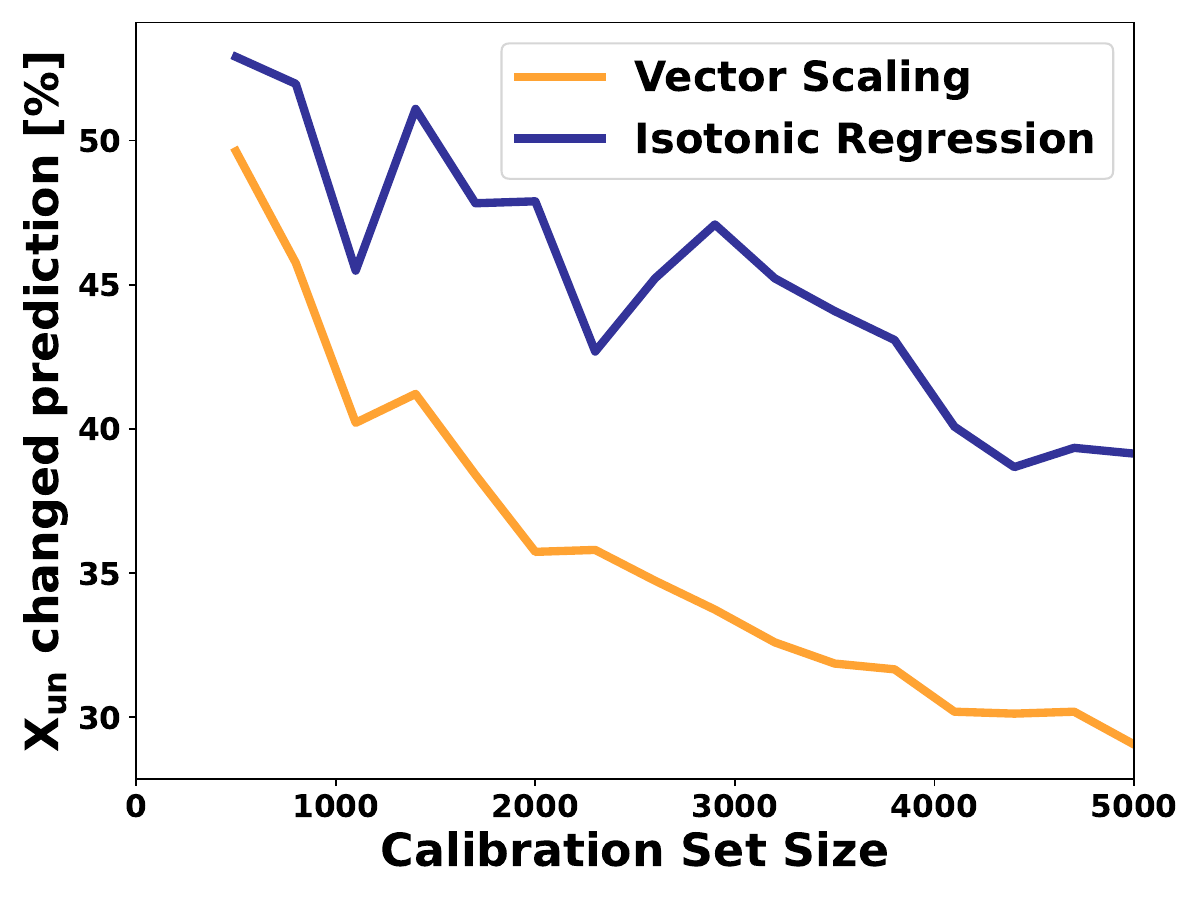}
    \label{fig:uncertain_por}}
    \caption{(a) Non-monotonic methods (e.g., vector scaling, isotonic regression) reduce accuracy, with the effect worsening as the calibration set shrinks. (b) Non-monotonic methods alter a large proportion of high-uncertainty predictions, with this proportion increasing as the calibration size decreases (ResNet-110 on CIFAR-100).}
    \label{fig:intro}
\end{figure}

Non-monotonic calibration maps can be easily parameterized and have demonstrated strong performance in some cases~\citep{rahimi_post-hoc_2022, kull_beyond_2019}. However, without enforcing monotonicity, and given that logits reside in a low-dimensional, abstract feature space derived from the original data, there is a risk of overfitting the calibration set logits. This overfitting can lead to a calibration map that fails to generalize well to testing sets. Moreover, since non-monotonic calibration maps do not preserve the ranking of predicted probabilities from the base model, they can negatively impact classification accuracy. To demonstrate this issue, we perform experiments with non-monotonic methods on CIFAR-100 with a wide range of calibration set sizes.  As shown in Figure~\ref{fig:acc_dec}, we observe a decrease in accuracy after non-monotonic calibration, which becomes more pronounced when the calibration set is smaller.

% more over although the change of sub one percent accuarcy may not look significant the porpotion thing

Additionally, changes in predictions or rankings predominantly occur in samples with high uncertainty— the cases where retaining the base model's ranking precision is crucial. Let $X_{un}$ denote the set of test samples where the top predicted probability from the base model is below 0.7, indicating high uncertainty. As shown in Figure~\ref{fig:uncertain_por}, non-monotonic methods alter the top prediction for a large proportion of samples in $X_{un}$, with this proportion increasing as the size of the calibration set decreases.

In our opinion, the primary role of post-hoc calibration is to correct systematic over- or under-confidence rather than to refine class rankings. By enforcing monotonicity, the calibration map adjusts probability estimates without altering the relative order of predictions. Thus, monotonicity can be viewed as a regularization constraint that enhances the robustness of the calibration map against a small or out-of-division calibration set.

While monotonicity is an important property for post-hoc calibration methods, how to perform monotonic calibration in a multiclass setting using a parameterized calibration map that is both interpretable and robust remains an open question. Existing monotonic calibration methods either rely on a small number of parameters, such as temperature scaling~\citep{guo_calibration_2017} and ensemble temperature scaling~\citep{zhang_mix-n-match_2020} or employ black-box neural network-based parameterized models that lack robustness and interpretability for the calibration map~\citep{rahimi_intra_2020, tomani_parameterized_2022}.

% The proposed method utilizes a constrained calibration map to transform logits in an interpretable and parameterized manner and

In this paper, we propose a family of novel monotonic post-hoc calibration maps that perform Instance-Wise Monotonic Calibration by Constrained Transformation (MCCT). By applying constrained transformations to sorted logits, the proposed method enables an interpretable and parameterized transformation while ensuring a robust and expressive calibration map that remains strictly monotonic. Implementing the calibration map through constrained optimization allows the model to learn a calibration map in an efficient and data-effective manner. Extensive experiments on CIFAR-10, CIFAR-100, and ImageNet demonstrate that our proposed methods, MCCT and its variant MCCT-I, outperform the existing state-of-the-art calibration methods. The source code is available at \url{https://github.com/YunruiZhang/Calibration-by-Constrained-Transformation}

% Another monotonicity order is for istance PX_i > Py_i the transformed Hx_i > Hy_i aka inter insatece ordering of probs. I am not sure if my proposed method is following this. I will check later and if it is. It will be another good selling point which I believe is quite important

\section{Preliminary}

Consider a multiclass classification problem where we have $\mathbf{x} \in \mathbb{R}^{d}$ representing the input features and the one-hot encoded label $y = (y_1, \ldots, y_m) \in \{0,1\}^{m}$, such that $\sum_{i=1}^{m} y_i = 1$, where $m$ is the number of classes.

Let $\mathcal{F}: \mathbb{R}^{d} \to \Delta^{m-1}$ be a probabilistic classifier that outputs the predicted class probability vector $\hat{p}(y|x) = (\hat{p}_{1}, \ldots, \hat{p}_m) \in [0,1]^{m}$ with $\sum_{i=1}^{m} \hat{p}_i = 1$ for each sample $x$. For deep neural networks, the class probability vector is usually generated by the softmax function: $\hat{p}_i = \frac{\exp(z_i)}{\sum_{j=1}^{m} \exp(z_j)}$, where $Z = (z_1, \ldots, z_m) \in \mathbb{R}^{m}$ are the logits generated by the neural network for each sample before the softmax operation.

% define the calibration notations and strength 
% define the calibration and introduce the ECE
\subsection{Calibration}
We want the predicted class probabilities to be calibrated for a classifier, meaning that the output probabilities should be close to the true probabilities of the corresponding classes. The strongest notion of calibration is complete calibration \citep{kull_beyond_2019}, defined as follows:

\begin{definition} \textbf{Complete Calibration}. A classifier $\mathcal{F}$ that outputs $\{\hat{p}(y_i|\mathbf{x})\}_{i=1}^m$ is complete calibrated if  $\hat{p}(y_i|\mathbf{x}) = p(y_{i}|\mathbf{x})$ for all classes $y_i$.
\label{CAL:1}
\end{definition}

Definition~\ref{CAL:1} represents the strictest form of calibration, where the predicted probabilities exactly match the true probabilities. A more relaxed notion is class-wise calibration \citep{zadrozny2002transforming}, defined as follows:

\begin{definition} \textbf{Class-wise Calibration}. A classifier $\mathcal{F}$ is classwise calibrated for a class $y_i$ if $\hat{p}(y_i|\mathbf{x}) = p(y_{i}|\mathbf{x})$.
\label{CAL:3}
\end{definition}
Another weaker but commonly used notion of calibration is confidence calibration \citep{guo_calibration_2017}, which requires only the top predicted probability (i.e., confidence) to be calibrated:

\begin{definition} \textbf{Confidence Calibration}.  A classifier $\mathcal{F}$ is confidence calibrated if $\argmax_y \hat{p}(y|\mathbf{x}) = \argmax_y p(y|\mathbf{x})$ and $\max \hat{p}(y|\mathbf{x}) = \max p(y|\mathbf{x})$. 
\label{CAL:2}
\end{definition}

% ranking based ECE top 1 top 2 etc, with the equal bin size instead of equal bin width
\subsection{Calibration Error}
While calibration notions are well-defined, measuring calibration error remains an open question. In practical applications, the ground-truth class probabilities are rarely available, making it impossible to directly measure the complete calibration error $\mathbb{E}_{X}[|\hat{p}(y|\mathbf{x}) - p(y|\mathbf{x})|]$, or even class-wise or top-label calibration error without access to true class probabilities.

A common approach for estimating calibration error is the Expected Calibration Error (ECE)~\citep{naeini2015obtaining, guo_calibration_2017}, which approximates the top-label confidence miscalibration. ECE partitions the samples based on their predicted confidence $\max \hat{p}(y|\mathbf{x})$ into $K$ equal-width bins. Each bin $B_k$ contains samples whose confidence falls within the interval $I_k = \left(\frac{k-1}{K}, \frac{k}{K}\right]$. The calibration error is then computed as the weighted average of the absolute difference between the bin's accuracy and average confidence.

\begin{equation}
    ECE = \sum\limits_{k=1}^{K}|acc(B_{k}) - conf(B_{k})|
    \label{eq:ECE}
\end{equation}
\[acc(B_{k}) = \frac{1}{|B_{k}|}\sum_{i \in B_{k}}\mathbf{1}(\hat{y_i} = y_i)\]
\[conf(B_{k}) = \frac{1}{|B_{k}|}\sum_{i \in B_{k}}\max \hat{p}(y|\mathbf{x_i})\]

Other variants of ECE have been proposed in previous work, such as Maximum Calibration Error~\citep{naeini2015obtaining}, Class-wise Calibration Error~\citep{kull_beyond_2019}, and adaptive ECE~\citep{nixon_measuring_2020}. However, ECE remains the most commonly used calibration measure due to its data efficiency and convenience for visualization.

% we use ECE becasue it is interpratable by reliablity diagram easy to use other methic have similar probmel relatively speakin gsample effcient

% define post hoc calibration 
% define the instance wise monotonic
\subsection{Post-hoc Calibration}
For the task of post-hoc calibration, we assume a set of unseen calibration data $\mathbf{X}_{c} \in \mathbb{R}^{d}$ with corresponding one-hot encoded labels $\mathbf{Y}_{c}$, probability outputs $\hat{P}_{c}$, and logits $Z_{c}$ provided by a deep neural network classifier $\mathcal{F}$.

The goal of post-hoc calibration is to learn a calibration map $T: \Delta^{m-1} \to \Delta^{m-1}$, trained on $\hat{P}_{c}$ and $\mathbf{Y}_{c}$, that transforms the model’s probabilistic outputs at test time to improve calibration. In the context of deep neural networks, the calibration map can also be learned in the logit space, where $T: \mathbb{R}^{m} \to \mathbb{R}^{m}$ is trained on $Z_{c}$ and $\mathbf{Y}_{c}$ to transform the testing time logits before softmax normalization. Logits reside in a less constrained feature space with more degrees of freedom than probability outputs. By operating on the logit space, we avoid the inherent constraints of the probability simplex and leverage the unconstrained logit geometry to improve calibration.

As outlined in the introduction, one desirable property for the calibration map is instance-wise monotonicity, meaning that the calibration transformation preserves the ordering of class probabilities or logits from the original classifier’s predictions.
\begin{definition} \textbf{Instance-wise monotonic calibration}.  
A calibration map \( T: \mathbb{R}^m \to \mathbb{R}^m \) is \textit{instance-wise monotonic} if for every logit vector or probability output \( Z \in \mathbb{R}^m \), \\ 

\centerline{$\forall i, j \in \{1, \ldots, m\}, \quad Z_i \geq Z_j \implies T(Z)_i \geq T(Z)_j.$}
 
\label{CAL:4}
\end{definition}
This property is also called accuracy-preserving \citep{zhang_mix-n-match_2020} and intra-order preserving \citep{rahimi_intra_2020}.

\section{Method}
In this section, we propose two monotonic calibration maps, MCCT and MCCT-I, based on the logit space and the corresponding implementation of the calibration map using constrained optimization. 

First, we define a sorting function $S: \mathbb{R}^{m} \to \mathbb{R}^{m}$ that sorts a given logit vector $Z$ into its sorted version $S(Z) = (Z_{(1)}, Z_{(2)}, \ldots, Z_{(m)})$, where \( Z_{(1)} < Z_{(2)} < \ldots < Z_{(m)} \). To prevent ties in the sorting process, our proposed calibration map also makes the following assumption on each sample's output logit:

\begin{assumption}
For every logit vector \( Z = (Z_1, \dots, Z_m) \in \mathbb{R}^m \), its components are pairwise distinct: \\

\centerline{$\forall i, j \in \{1, \ldots, m\},\ (i \neq j) \implies Z_i \neq Z_j.$}
\label{ass1}
\end{assumption}

Assumption \ref{ass1} ensures strict total ordering of logits, avoiding ties in class rankings by assuming no two logits have the same value. This assumption is reasonable for logits produced by a deep neural network, as the logit space is an unconstrained and continuous feature space $ \mathbb{R}^m$ with the number of classes $m$ usually relatively small. Empirically, in our experiments across various neural networks and datasets, we have not encountered tied logits in any sample, even for the more extreme cases such as ImageNet-1K, which has 1000 classes. 

\subsection{MCCT}
Now we describe the proposed monotonic calibration map based on the logit space:

\begin{theorem}
For transformation \( f: \mathbb{R}^m \to \mathbb{R}^m \), defined as  
\[
f(Z) = S^{-1}\big(S(Z) \odot w + b\big),
\]  
is monotonic if: \\  
- \( w \in \mathbb{R}^m \) vector satisfy \( w_1 \leq w_2 \leq \ldots \leq w_m \) and \(w_{i} > 0\) \\  
- \( b \in \mathbb{R}^m \) vector satisfy \( b_1 \leq b_2 \leq \ldots \leq b_m \) \\ 
- \( S^{-1} \) is the inverse sorting operator that maps the sorted-and-transformed logit vector back to the original logit $Z$ indices.
\label{the1}
\end{theorem}

The proof for Theorem \ref{the1} is intuitive, since $(S(Z)$ and also $w$ and $b$ are non-descending, the transformed sorted vector $\hat{Z} = S(Z) \odot w + b$  preserves the ordering of $S(Z)$. So $\forall i,j \in \{1\ldots m\}$ if $Z_i < Z_j$ then $\hat{Z}_i < \hat{Z}_j$. Since the invert sorting operator $S^{-1}$ maps $\hat{Z}$ back to the original logit indices by reversing the sorting permutation of $S$, it preserves this inequality in the original indices. Thus proving monotonicity. 

The proposed calibration map offers an intuitive way to parameterize the calibration transformation in the logit space while preserving the original order of predictions. Moreover, it provides interpretability, as the effects of the transformation can be directly understood through the applied scaling and bias in the sorted logit domain. By sorting the logits we could connect the parameters to ranking instead of the classes which improves the sample efficient in the cases of many classes or small calibration set.

We employ a constrained optimization approach to implement the proposed calibration map in Theorem~\ref{the1} and learn the parameters $w$ and $b$. Specifically, we enforce monotonicity by requiring the consecutive differences of \(w\) and \(b\) to be positive. Formally, we define the difference vectors \(d_w \in \mathbb{R}^{m-1}\) and \(d_b \in \mathbb{R}^{m-1}\) for the weight vector \(w \in \mathbb{R}^{m}\) and the bias vector \(b \in \mathbb{R}^{m}\) as :
\[
\forall i \in \{1 \ldots m-1\} \quad d_{w,i} = w_{i+1} - w_i, \quad d_{b,i} = b_{i+1} - b_i
\]
Enforcing the $d_{w}$ and $d_{b}$ to be strictly positive in the optimization process guarantees that both \(w\) and \(b\) are non-decreasing, thereby preserving the order of the logits after transformation. The parameters are learned by minimizing the negative log-likelihood (NLL) loss between the softmax of the transformed logits \(\hat{Z}\) and the true labels. Formally, we minimize the following loss to learn the $w$ and $b$ parameters:
\[
\min_{w,b} \quad \mathcal{L}_{\text{NLL}}\big(\text{softmax}(S^{-1}(S(Z) \odot w + b)), y\big)
\]
subject to:
\[
\forall i \in \{ 1 \ldots m-1 \}, \quad d_{w,i} \geq 0, \quad d_{b,i} \geq 0
\]
Where the gradient of parameter $w$ and $b$ for each sample are 
% \[
% \frac{\partial{\mathcal{L}}}{\partial{w}} = Z \odot (\hat{p}_c-y),
% \]
% \[
% \frac{\partial{\mathcal{L}}}{\partial{b}} = \hat{p}_c-y
% \]

\[
\frac{\partial{\mathcal{L}}}{\partial{w}} = S(Z) \odot (\hat{p}_c-y), \quad \frac{\partial{\mathcal{L}}}{\partial{b}} = \hat{p}_c-y
\]

where $\hat{p}_c =\text{softmax}(S^{-1}(S(Z) \odot w + b))$, the probability output after the softmax operator.

We choose not to apply any additional normalization to the logits during post-hoc calibration, as it is standard practice in modern deep neural network training and inference to normalize inputs using the mean and standard deviation computed from the training set. This preprocessing step regularizes the scale and distribution of the logits. Empirically, we observe that for each logit rank, the resulting distribution is approximately normal.

\subsection{MCCT-I}
% As we can see from the gradient of $w$ of MCCI it indirectly relatied to the $w$ parameter
We also propose an alternative calibration map as a variant of Theorem~\ref{the1}, where the primary difference is that $w$ is an inverse scaling factor in the transformation. This calibration map is defined as:
\begin{theorem}
For transformation \( f: \mathbb{R}^m \to \mathbb{R}^m \), defined as  
\[
f(Z) = S^{-1}\left(\frac{S(Z)}{w} + b\right),
\]  
is monotonic if: \\  
- \( w \in \mathbb{R}^m \) vector satisfy \( w_1 \geq w_2 \geq \ldots \geq w_m \) and \(w_{i} > 0\)\\  
- \( b \in \mathbb{R}^m \) vector satisfy \( b_1 \leq b_2 \leq \ldots \leq b_m \) \\ 
- \( S^{-1} \) is the inverse sorting operator that maps the sorted-and-transformed logit vector back to the original logit $Z$ indices.
\label{the2}
\end{theorem}

Theorem~\ref{the2} is a variant of Theorem~\ref{the1}, with the only difference being that \(w\) acts as an inverse scaling factor. It is straightforward to see that the transformation map described in Theorem~\ref{the2} is mathematically equivalent to that of Theorem~\ref{the1}, differing only in how the constraint on \(w\) is formulated. The key distinction between these two calibration maps lies in the optimization process, where the implementation remains similar, but the parameters have different gradients. The implementation of Theorem~\ref{the2} is defined as:
\[\min_{w,b} \quad \mathcal{L}_{\text{NLL}}\left(\text{softmax}\left(S^{-1}\left(\frac{S(Z)}{w} + b\right)\right), y\right)\]
subject to:
\[\forall i \in \{ 1 \ldots m-1 \}, \quad g_{w,i} \geq 0, \quad g_{b,i} \geq 0\]
\[\forall j \in \{1\ldots m\}, w_j > 0\]

Where the constraint terms vector $g_w$ and $g_b$ are defined as:
\[g_{w,i} = w_{i} - w_{i+1}, \quad g_{b,i} = b_{i+1} - b_i, \quad \forall i \in \{1 \ldots m-1\}\]
The gradient for parameter $w$ and $b$ for each sample are
% \[
% \frac{\partial{\mathcal{L}}}{\partial{w}} = -\frac{Z}{w^2} \odot (\hat{p}_c-y),
% \]
% \[\frac{\partial{\mathcal{L}}}{\partial{b}} = \hat{p}_c-y
% \]
\[
\frac{\partial{\mathcal{L}}}{\partial{w}} = -\frac{S(Z)}{w^2} \odot (\hat{p}_c-y), \quad \frac{\partial{\mathcal{L}}}{\partial{b}} = \hat{p}_c-y
\]

The gradient for the calibration map in Theorem~\ref{the2} is scaled by $\frac{1}{w^2}$ in comparison with Theorem~\ref{the1}, which acts as an implicit regularization discouraging large values of $w$. Empirically, this implicit regularization has benefited datasets with many classes, such as ImageNet-1K, where overconfidence is less pronounced compared to datasets with fewer classes. In such cases, enforcing smaller $w$ through implicit regularization improves calibration performance.

\subsection{Scalability in Large Number of Class}\label{Sec:scal}

% Empirical results on ImageNet-1K, which has 1,000 classes and uses a calibration set of 25K samples, show that the optimization completes within 55 minutes on a single-threaded CPU, which we believe is a reasonable computational cost for fitting a calibration map.

For multiclass calibration, a significant challenge is handling many classes. The proposed calibration methods, MCCT and MCCT-I, scale well to problems with many classes since the number of parameters grows linearly with the number of classes. Nonetheless, it is important to consider extreme cases with an even more significant number of classes.

A reasonable assumption about the probabilistic output of a classifier is that, as the number of classes increases, the output remains constrained to the probability simplex $\Delta^{m-1}$, where all class probabilities sum to one. In such cases, the lower-ranked probabilities, and consequently their corresponding logits, may become negligible and contain little helpful information for calibration. Thus, for extremely large-scale classification tasks, it may be beneficial to discard the lowest-ranked logits to reduce computational complexity while preserving calibration effectiveness.

For MCCT and MCCT-I, discarding the lowest-ranked logits can be implemented by truncating $S(Z)$ and the corresponding label $y$ only to consider the top $k$ logits, effectively reducing the number of parameters to $k$ for $w$ and $b$. Specifically, we define the reduced sorted logit vector as $ S(Z)^k = S(Z)_i \quad \forall i \in \{m-k, \dots, m\},$ with the corresponding reduced label vector $ y^k $ indexed accordingly.  

During inference, for the lower ranked $ S(Z)_i \quad \forall i \in \{1, \dots, m-k\}$ logits, the transformation is carried out by using $ w_1 $ and $ b_1 $, the first elements of the weight and bias vectors, respectively, to scale the non-top-k logits.

\section{related work}

Numerous post-hoc calibration methods have been explored in the literature. As discussed in the previous section, a key distinction among these methods is whether they enforce monotonicity. Non-monotonic parametric methods include vector scaling, matrix scaling \citep{guo_calibration_2017}, and Dirichlet scaling \citep{kull_beyond_2019}. However, since they are heavily parametrized in an unconstrained manner, these methods are not data-efficient, and their lack of monotonicity becomes more severe as the calibration set size decreases \citep{rahimi_intra_2020}.

\textbf{Monotonic post-hoc calibration.} Temperature scaling \citep{guo_calibration_2017, platt1999probabilistic, hinton2015distilling} is the simplest post-hoc calibration method that enforces monotonicity. It uses a single parameter to scale the output logits for calibration. Later work, such as Ensemble Temperature Scaling \citep{zhang_mix-n-match_2020}, extends temperature scaling by incorporating three additional weight parameters to improve its expressivity. More recently, there have been attempts to perform monotonic calibration using more parameterized approaches by learning a neural network for calibration maps. Intra-Order Preserving Calibration \citep{rahimi_intra_2020} employs a monotonic neural network to transform individual logits for improved calibration. Additionally, parameterized temperature scaling utilizes a nonlinear neural network to select different temperature values for each sample, assuming that different samples have different levels of over or underconfidence and need different temperature parameters for calibration.

\textbf{Non-parametric post-hoc calibration.} Apart from parametric approaches, many non-parametric methods have also been proposed, with binning-based approaches being the most popular category. These methods divide confidence outputs into mutually exclusive bins and assign a calibrated score to each bin. A notable example is histogram binning \citep{HB}. Several variants of binning have also been introduced, such as Bayesian Binning into Quantiles (BBQ) \citep{naeini2015obtaining}, Platt Binner Marginal Calibrator \citep{kumar_verified_2019}, and other closely related non-parametric methods like isotonic regression \citep{zadrozny2002transforming} and spline calibration \citep{gupta_calibration_2021}.  Similar to non-monotonic parametric methods, most non-parametric calibration methods suffer from data inefficiency and increased monotonicity violations as the size of the calibration set decreases.

\textbf{Training calibrated models.} Besides post-hoc calibration methods, several approaches focus on training models that are well-calibrated. Techniques such as label smoothing \citep{muller_when_2020}, focal loss \citep{mukhoti2020calibrating}, dual focal loss \citep{dual_focal_loss}, logit normalization \citep{wei2022mitigating} have been shown to improve the calibration of deep neural networks.

\textbf{Calibration Metric.} The expected calibration error (ECE) \citep{naeini2015obtaining} is the most commonly used calibration metric in the literature. Its popularity stems from its data efficiency and close relationship with reliability diagrams, making it both interpretable and easily visualized. However, one of its key weaknesses is its dependence on the number of bins $b$, which introduces sensitivity to the choice of this hyperparameter. Various variants of ECE have been proposed to mitigate this issue \citep{kumar_verified_2019, nixon_measuring_2020, ece_flaw}, but still rely on alternative assumptions or additional hyperparameters. In our evaluation, in addition to the standard ECE, we use a variant of the adaptive ECE proposed by \citet{nixon_measuring_2020}, which employs equal-sized binning rather than equal-interval binning based on the top-label confidence. The error is computed as the sum of the un-normalized differences between accuracy and average confidence within each bin. Beyond confidence calibration, other notions of calibration error have been explored, such as class-wise ECE. However, these approaches tend to be less data-efficient and often exhibit inconsistencies with other calibration metrics \citep{tomani_parameterized_2022, nixon_measuring_2020}. ECE-KDE \citep{zhang_mix-n-match_2020} is another variant of ECE that replaces histogram-based binning with a non-parametric density estimator for calibration error estimation, implemented using kernel density estimation (KDE). Where the bandwidth is set as $h = 1.06 \hat{\sigma} n_e^{-1/5}$, where $\hat{\sigma}$ is the standard deviation of the confidence scores, following a popular bandwidth selection method \citep{scott2015multivariate}. 

\section{Experiments}
We evaluates the proposed calibration methods, MCCT and MCCT-I, on diverse deep-learning models across different datasets and compared their performance with the existing state-of-the-art post-hoc calibration methods.

\subsection{Datasets and Base Classifier}

We perform our experiments on three computer vision datasets: CIFAR-10 \citep{krizhevsky2009learning}, CIFAR-100 \citep{krizhevsky2009learning}, and ImageNet-1K \citep{5206848} with the number of class ranging from $10$ to $1000$. For all datasets, we assess calibration performance across different neural network architectures. For CIFAR-10/100, we use a calibration set of 5,000 samples and a test set of 10,000 samples. For ImageNet-1K, we use a validation set of 25,000 samples and a test set of 25,000 samples.

As for neural network classifier model, for CIFAR-10, we use LeNet-5 \citep{lenet}, DenseNet-40 \citep{densenet}, and ResNet-110 \citep{he2016deep}. For CIFAR-100, we use Wide-ResNet-32 \citep{zagoruyko2016wide} and ResNet-110 \citep{he2016deep}. For ImageNet-1K, we use ResNet-50/152 \citep{he2016deep}, ViT-B/16 \citep{Vit}, Inception-V3 \citep{inception}, and Wide-ResNet-50-2 \citep{zagoruyko2016wide}.  

\subsection{Baselines}

To compare the calibration performance of MCCT and MCCT-I, we select the following seven baseline methods: Temperature Scaling (TS) \citep{guo_calibration_2017}, Vector Scaling (VS) \citep{guo_calibration_2017}, Histogram Binning (HB) \citep{HB}, Ensemble Temperature Scaling with Negative Log-Likelihood Loss (ETS) \citep{zhang_mix-n-match_2020}, Ensemble Temperature Scaling with Mean Squared Error Loss (ETS-MSE) \citep{zhang_mix-n-match_2020}, Parameterized Temperature Scaling (PTS) \citep{tomani_parameterized_2022}, Intra-Order Preserving Calibration (DIAG) \citep{rahimi_intra_2020}.

% \begin{description}
%     \item Temperature Scaling (TS) \citep{guo_calibration_2017}  
%     \item Vector Scaling (VS) \citep{guo_calibration_2017}  
%     \item Histogram Binning (HB) \citep{HB}  
%     \item Ensemble Temperature Scaling with Negative Log-Likelihood Loss (ETS) \citep{zhang_mix-n-match_2020}  
%     \item Ensemble Temperature Scaling with Mean Squared Error Loss (ETS-MSE) \citep{zhang_mix-n-match_2020}  
%     \item Parameterized Temperature Scaling (PTS) \citep{tomani_parameterized_2022}  
%     \item Intra-Order Preserving Calibration (DIAG) \citep{rahimi_intra_2020}  
% \end{description}
We also include the original uncalibrated classifier result (Uncalibrated) for comparison. Among all the baselines, the VS and HB are not monotonic and do not preserve the accuracy or the prediction ranking in the calibration process.

\subsection{Results}

\begin{table*}[ht]
\centering
\begin{tabular}{ccccccccccc}
\textbf{Model} &
  \textbf{Uncalibrated} &
  \textbf{TS} &
  \textbf{VS} &
  \textbf{HB} &
  \textbf{ETS} &
  \textbf{ETS MSE} &
  \textbf{PTS} &
  \textbf{DIAG} &
  \textbf{MCCT} &
  \textbf{MCCT-I} \\
  \hline
  \multicolumn{11}{c}{CIFAR10} \\
  \hline
Lenet5          & 5.18  & 1.67 & 1.28  & 4.67 & 1.61 & 1.49 & 11.48 & 1.42 & \textbf{1.19} & \textbf{1.19} \\
Densenet 40      & 5.50  & \textbf{0.95} & 0.98 & 1.61 & 1.17 & 1.09 & 3.96  & 1.05 & 1.06 & 1.06 \\
W-Resnet 32  & 4.51  & 0.78 & 0.85  & 0.73 & 0.68 & 0.69 & 3.49  & 0.73 & \textbf{0.54} & \textbf{0.54} \\
Resnet 110       & 4.75  & 1.13 & 1.30  & 1.13 & 0.82 & \textbf{0.58} & 4.26  & 0.79 & 0.76 & 0.76 \\
    \hline
    \multicolumn{11}{c}{CIFAR100} \\
    \hline
W-Resnet 32 & 18.78 & 1.47 & 1.58 & 8.43 & 1.32 & 1.25 & 5.82  & 2.00 & \textbf{1.06} & \textbf{1.11} \\
Resnet 110      & 18.48 & 2.38 & 2.54 & 8.53 & 1.48 & 1.68 & 7.15  & 1.50 & \textbf{1.28} & \textbf{1.32} \\
    \hline
    \multicolumn{11}{c}{ImageNet} \\
    \hline
Resnet 50             & 3.66  & 2.25 & 1.77 & 7.32 & 1.39 & 1.35 & 1.09  & 1.81 & 0.99 & \textbf{0.95} \\
Resnet 152            & 4.78  & 2.02 & 1.86 & 7.73 & 1.10 & \textbf{1.07} & 1.22  & 1.69 & 1.28 & 1.18 \\
ViT-B/16              & 5.86  & 3.72 & 4.23 & 7.04 & 3.36 & 3.10 & 2.50  & \textbf{0.67} & 1.79 & 1.79 \\
Inception v3        & 18.27 & 5.41 & 27.93 & 7.61 & 2.94 & 2.52 & 1.62  & 0.98 & 0.72 & \textbf{0.61} \\
W-ResNet 50    & 5.43  & 2.95 & 3.11 & 7.70 & 1.77 & 1.56 & 1.16  & 1.14 & 1.13 & \textbf{1.08} \\
\hline
Average rank & 9.45 & 6.32 & 6.64 & 8.55 & 4.82 & 4.00 & 6.55 & 4.32 & 2.32 & \textbf{2.05} \\
\hline
\end{tabular}
\caption{ECE (with the number of bins set to 15, smaller is better) on benchmark datasets and models with different calibration methods. Calibration methods VS and HB are not monotonic and suffer from changes in accuracy and prediction rankings. The reported numbers are averaged over 10 runs and are in the scale of $10^{-2}$.}

\label{Tab1}
\end{table*}

\begin{table*}[ht]
\centering
\begin{tabular}{ccccccccccc}
\textbf{Model} &
  \textbf{Uncalibrated} &
  \textbf{TS} &
  \textbf{VS} &
  \textbf{HB} &
  \textbf{ETS} &
  \textbf{ETS MSE} &
  \textbf{PTS} &
  \textbf{DIAG} &
  \textbf{MCCT} &
  \textbf{MCCT-I} \\
  \hline
  \multicolumn{11}{c}{CIFAR10} \\
  \hline
Lenet5           & 4.66  & 1.26 & 1.19  & 3.68 & 1.22 & 1.22 & 11.01 & 1.10 & \textbf{0.88} & \textbf{0.88} \\
Densenet 40       & 5.01  & \textbf{1.14} & 1.25  & 2.16 & 1.31 & 1.35 & 3.58  & 1.31 & 1.30 & 1.30 \\
W-Resnet 32   & 4.06  & 1.16 & 1.28  & 1.44 & 1.16 & 1.17 & 3.26  & 1.35 & \textbf{1.04} & \textbf{1.04} \\
Resnet 110       & 4.34  & 1.24 & 1.41  & 1.66 & 1.25 & \textbf{1.08} & 3.94  & 1.23 & \textbf{1.11} & \textbf{1.11} \\
    \hline
    \multicolumn{11}{c}{CIFAR100} \\
    \hline
W-Resnet 32  & 18.16 & \textbf{1.09} & 1.17  & 7.75 & 1.25 & 1.33 & 5.63  & 2.11 & 1.13 & \textbf{1.12} \\
Resnet 110       & 17.82 & 1.71 & 1.91  & 8.94 & 1.45 & 1.47 & 6.90  & 1.69 & \textbf{1.17} & \textbf{1.18} \\
    \hline
    \multicolumn{11}{c}{ImageNet} \\
    \hline
Resnet 50            & 3.02  & 1.85 & 1.29  & 7.99 & 1.44 & 1.57 & \textbf{1.00}  & 2.05 & 1.07 & \textbf{1.00} \\
Resnet 152            & 4.24  & 1.61 & 1.40  & 7.40 & 1.32 & 1.38 & \textbf{1.13}  & 1.95 & 1.23 & 1.21 \\
ViT-B/16              & 5.78  & 3.79 & 4.09  & 6.68 & 3.25 & 2.97 & 2.40  & \textbf{0.92} & 1.74 & 1.75 \\
Inception v3         & 17.71 & 5.05 & 27.60 & 8.30 & 2.88 & 2.33 & 1.39  & 1.17 & \textbf{0.67} & \textbf{0.68} \\
W-ResNet 50     & 4.83  & 2.42 & 2.57  & 7.46 & 1.78 & 1.79 & \textbf{0.94}  & 1.34 & 1.07 & 1.02 \\
\hline
Average rank & 9.45 & 5.32 & 6.00 & 8.90 & 4.95 & 5.05 & 5.86 & 5.05 & 2.27 & \textbf{2.14} \\
\hline
\end{tabular}
\caption{ECE-KDE on benchmark datasets and models with different calibration methods. Calibration methods VS and HB are not monotonic and suffer from changes in accuracy and prediction rankings. The reported numbers are averaged over 10 runs and are in the scale of $10^{-2}$.}

\label{Tab2}
\end{table*}

% ECE eq bin need some intro later
\begin{table*}[]
\begin{tabular}{ccccccccccc}
\textbf{Model} &
  \textbf{Uncalibrated} &
  \textbf{TS} &
  \textbf{VS} &
  \textbf{HB} &
  \textbf{ETS} &
  \textbf{ETS MSE} &
  \textbf{PTS} &
  \textbf{DIAG} &
  \textbf{MCCT} &
  \textbf{MCCT-I} \\
  \hline
  \multicolumn{11}{c}{CIFAR10} \\
  \hline
Lenet5         & 5.04  & 1.48          & 1.03  & 4.19  & 1.31 & 1.30          & 11.46         & 1.09          & \textbf{0.82} & \textbf{0.82} \\
Densenet40      & 5.49  & \textbf{0.92} & 0.93  & 2.67  & 1.19 & 1.32          & 3.95          & 1.10          & 1.09          & 1.09          \\
W-Resnet 32  & 4.48  & 0.70          & 0.56  & 2.84  & 0.69 & 0.71          & 3.40          & 0.52          & \textbf{0.42} & \textbf{0.42} \\
Resnet 110       & 4.75  & 0.93          & 1.13  & 2.29  & 0.87 & \textbf{0.42} & 4.26          & 0.92          & 0.79          & 0.79          \\
\hline
    \multicolumn{11}{c}{CIFAR100} \\
\hline
W-Resnet 32 & 18.78 & 1.17          & 1.33  & 8.21  & 1.35 & 1.45          & 5.79          & 1.86          & \textbf{0.96} & \textbf{0.97} \\
Resnet 110     & 18.48 & 2.09          & 2.25  & 9.65  & 1.36 & 1.47          & 7.14          & 1.42          & \textbf{1.07} & \textbf{1.06} \\
\hline
    \multicolumn{11}{c}{ImageNet} \\
\hline
Resnet 50             & 8.65  & 5.60          & 4.52  & 22.41 & 3.95 & 4.17          & 2.88          & 4.68          & \textbf{2.56} & \textbf{2.43} \\
Resnet 152            & 11.84 & 5.20          & 4.54  & 21.69 & 3.95 & 4.04          & \textbf{3.10} & 4.43          & 3.37          & 3.20          \\
ViT-B/16               & 14.70 & 10.51         & 10.57 & 18.43 & 8.24 & 7.54          & 6.32          & \textbf{2.25} & 4.71          & 4.72          \\
Inception v3        & 45.68 & 13.44         & 69.82 & 24.41 & 8.24 & 6.71          & 4.07          & 2.79          & \textbf{2.23} & 2.40          \\
W-ResNet 50    & 13.45 & 7.23          & 7.77  & 22.70 & 5.02 & 4.82          & 2.97          & 3.03          & 3.05          & \textbf{2.92}\\
\hline
Average rank & 9.45 & 6.00 & 6.00 & 8.91 & 5.00 & 5.09 & 6.09 & 4.36 & 2.18 & \textbf{1.91} \\
\hline
\end{tabular}
\caption{EQ-BIN ECE on benchmark datasets and models with different calibration methods. Calibration methods VS and HB are not monotonic and suffer from changes in accuracy and prediction rankings. The reported numbers are averaged over 10 runs and are in the scale of $10^{-1}$.}
\label{Tab3}
\end{table*}

Table \ref{Tab1} summarizes the results of our proposed calibration methods compared to the baselines in terms of ECE and average ranking, calculated across all datasets and base classifiers. Overall, MCCT and MCCT-I achieve the highest rankings among all methods, with MCCT-I slightly outperforming MCCT. Additionally, MCCT and MCCT-I achieve the highest number of best-performing results, ranking first in 7 out of 11 cases. Even in rare instances where MCCT or MCCT-I is not the best-performing calibration method, they remain competitive, typically ranking second and not far behind in terms of ECE. The performance of MCCT and MCCT-I is similar on datasets with fewer classes. However, as the number of classes increases, MCCT-I begins to outperform MCCT due to the implicit regularization effect introduced by the different gradients on parameter $w$, as explained in the previous section. Table \ref{Tab2} presents the results using the ECE-KDE metric \citep{zhang_mix-n-match_2020}, which remain mostly consistent with and correlate closely to ECE. Table \ref{Tab3} presents the results for Equal Bin ECE (EQ-BIN ECE), a variant of ECE that, instead of dividing samples by grouping the top-confidence predictions into bins of equal confidence intervals. The results are consistent with those of ECE and ECE-KDE, with MCCT and MCCT-I remaining the top-performing calibration methods by a significant margin.

Similar to TS, MCCT, and MCCT-I are interpretable since the calibration map is directly parameterized by intuitive parameters. By simply examining the $w$ and $b$ parameters, we can assess the degree of overconfidence and underconfidence in the model, as well as how these effects vary across probability rankings. In contrast, neural network-based approaches such as PTS and DIAG lack this level of interpretability.

Another observation from our experiments is that MCCT and MCCT-I are not sensitive to random seeds or initialization. In our experiments, the random seed has no impact on the results for MCCT and MCCT-I. In contrast, neural network-based methods such as PTS and DIAG exhibit high variance in ECE when trained with different random seeds. 

For initialization, MCCT sets the weight vector $w$ by assigning it $m$ equally spaced values in the interval $[0, 1]$, and $b$ vector is initialized to $0$. For MCCT-I, the $w$ vector is initialized to ones, and $b$ is initialized to zeros. Our experiments found that neither MCCT nor MCCT-I are sensitive to parameter initialization when optimized by sequential least squares programming (SLSQP) optimizer \citep{kraft1988software}.

\subsection{Data Efficiency}

% A crucial aspect of a calibration method is its data efficiency, aka how the post hoc calibration method perform when the calibration map is small. Since in many real-world applications, a large set set a side unseen validation set is not possible. In Figure \ref{fig:acc_dec}, we have already shown that for non-monotonic method the smaller calibrtion set would woresen the monotonicity violations. While for monotonic post-hoc calibration method, it is essential to maintain the calibration perfromce when the calibration set is small.
A crucial aspect of a calibration method is its data efficiency, i.e., how well the post-hoc calibration method performs when the calibration set is small. In many real-world applications, setting aside a large unseen validation set is not feasible. As shown in Figure \ref{fig:acc_dec}, a smaller calibration set increases monotonicity violations for non-monotonic methods. For monotonic post-hoc calibration methods, it is essential to maintain calibration performance even when the calibration set is limited.

\begin{figure}[h]
    \centering
    \subfloat[ImageNet]{\includegraphics[width= .49\columnwidth]{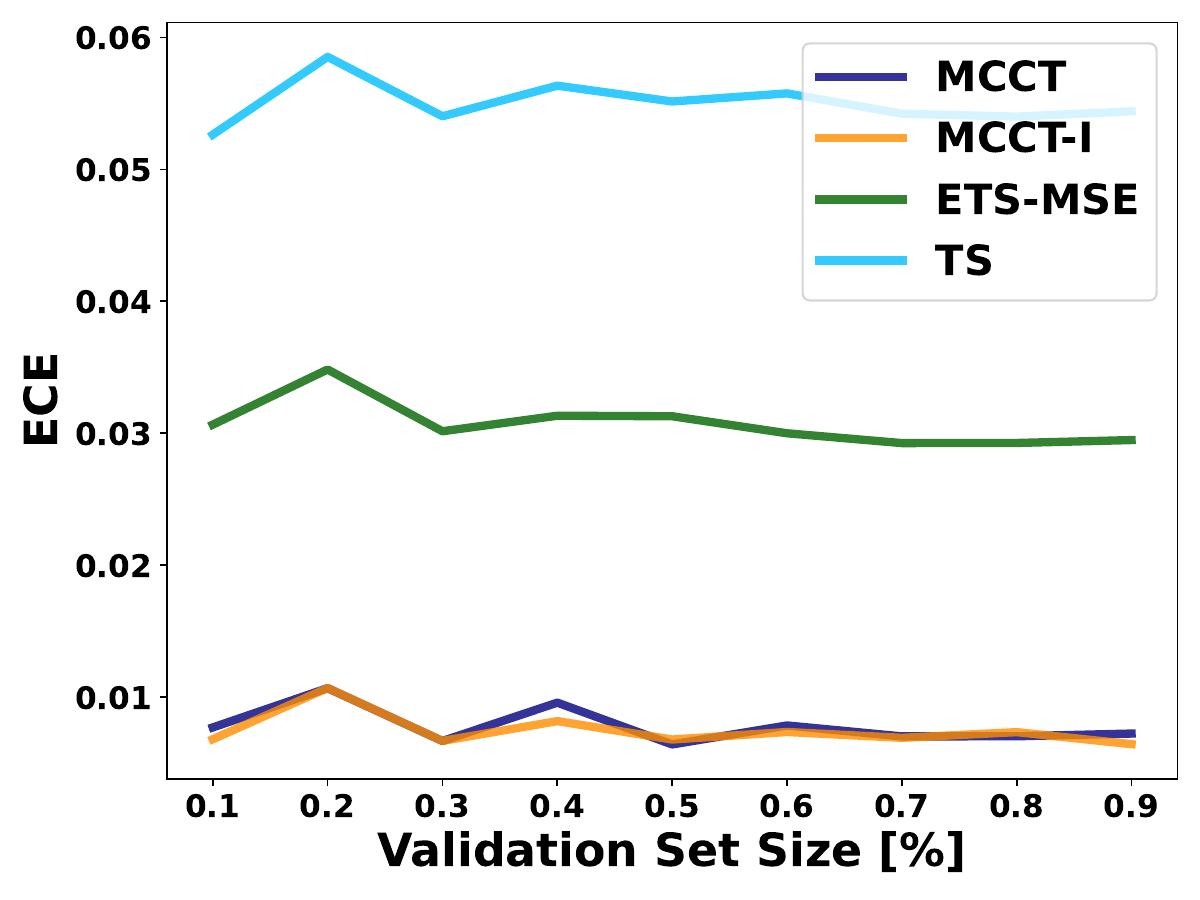}}
    \label{fig:eff_a}
    \subfloat[CIFAR-100]{\includegraphics[width= .49\columnwidth]{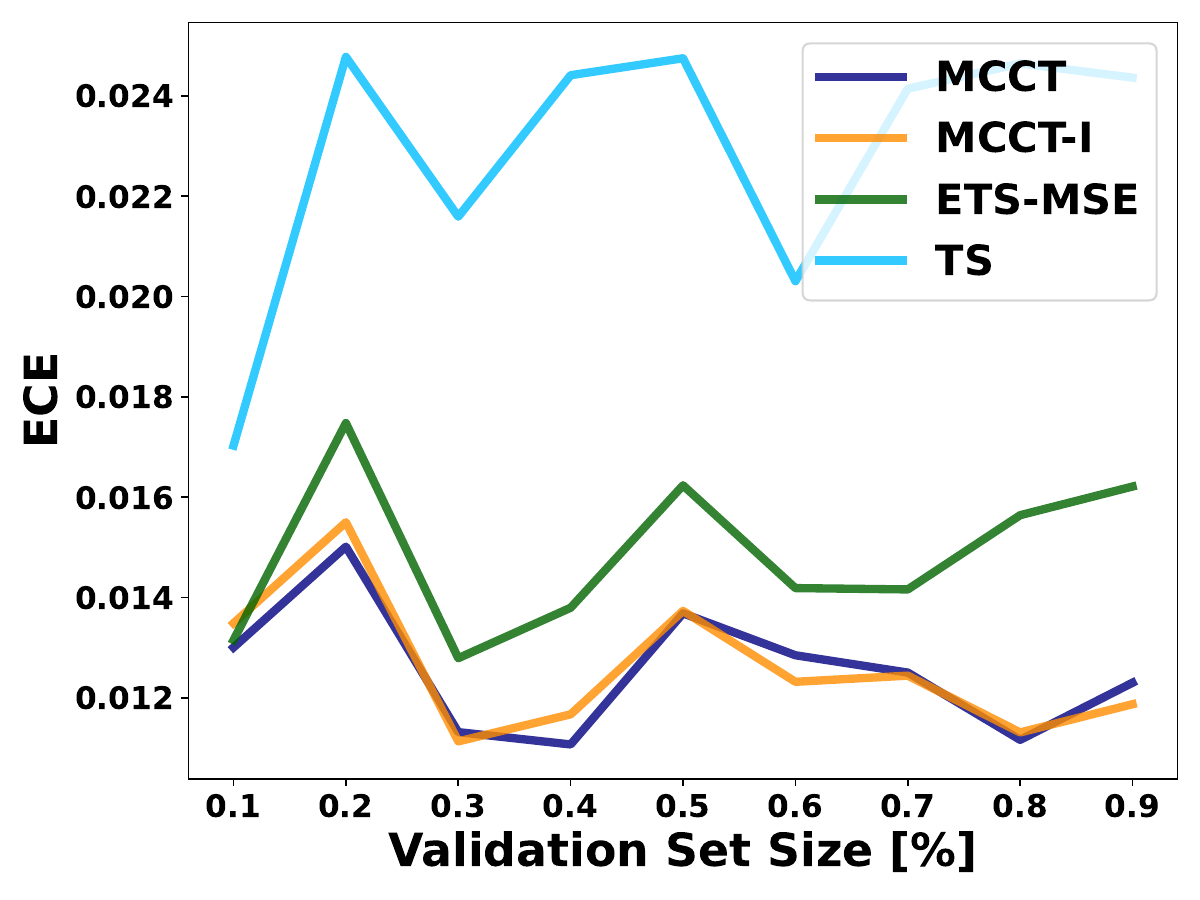}}
    \label{fig:eff_b}
    \caption{The impact of calibration set size on ECE for MCCT and MCCT-I compared to data-efficient methods TS and ETS. (a) ImageNet with InceptionNetV3. (b) CIFAR-100 with ResNet110.}
    \label{fig:eff_test}
\end{figure}

Calibration methods with fewer parameters, such as TS and ETS, have a notable advantage in data efficiency since they are less prone to overfitting on small calibration sets. To assess the data efficiency of MCCT and MCCT-I, we conduct experiments on ImageNet and CIFAR-100 using InceptionNetV3 and ResNet110 as base models. We fit MCCT and MCCT-I on varying calibration set sizes, ranging from $10\%$ to $90\%$ of the original size. We compare their performance with data-efficient methods such as TS and ETS. The results are presented in Figure \ref{fig:eff_test}.

As shown in Figure \ref{fig:eff_test}, MCCT and MCCT-I exhibit strong data efficiency, with no significant increase in ECE for ImageNet even when the calibration set size is reduced to as low as $10\%$ of its original size. For CIFAR-100, the ECE fluctuates by less than $30\%$ across different calibration set sizes, following a similar trend observed in TS and ETS. Moreover, MCCT and MCCT-I consistently outperform TS and ETS-MSE across all calibration set sizes. A potential reason for MCCT and MCCT-I performing well in low-data regimes is that the constraints on the transformation parameters act as a form of regularization, restricting the parameter search space and improving generalization.
% maybe add some anaylisis on this. The constrain act like a regulization? etc

\subsection{ImageNet top-k logits}
As discussed in Section~\ref{Sec:scal}, scalability can become a concern when handling an extremely large number of classes. In our experiments, both MCCT and MCCT-I complete training within a reasonable time on a single-threaded CPU. For CIFAR-10 and CIFAR-100, training time is negligible requiring less than 2 seconds and 6 seconds, respectively. For larger-scale datasets such as ImageNet, both methods finish training in under 55 minutes. While this training time is acceptable and does not render the method impractical, it may become a limitation as the number of classes increases.

To evaluate the top-$k$ logits selection method introduced in Section \ref{Sec:scal}, which aims to improve efficiency on datasets with a large number of classes, we conduct experiments on ImageNet-1K using ResNet-50 and ResNet-152 as base models. We explore a wide range of top-$k$ parameter values, varying $k$ from 100 to 1000. Our goal is to investigate the impact of top-$k$ selection on ECE, determining how many logits can be trimmed before significantly affecting calibration performance. Additionally, we analyze the computational efficiency gains achieved by reducing the number of selected logits.

\begin{figure}[htbp]
    \centering
    \subfloat[ResNet50]{\includegraphics[width= .5\columnwidth]{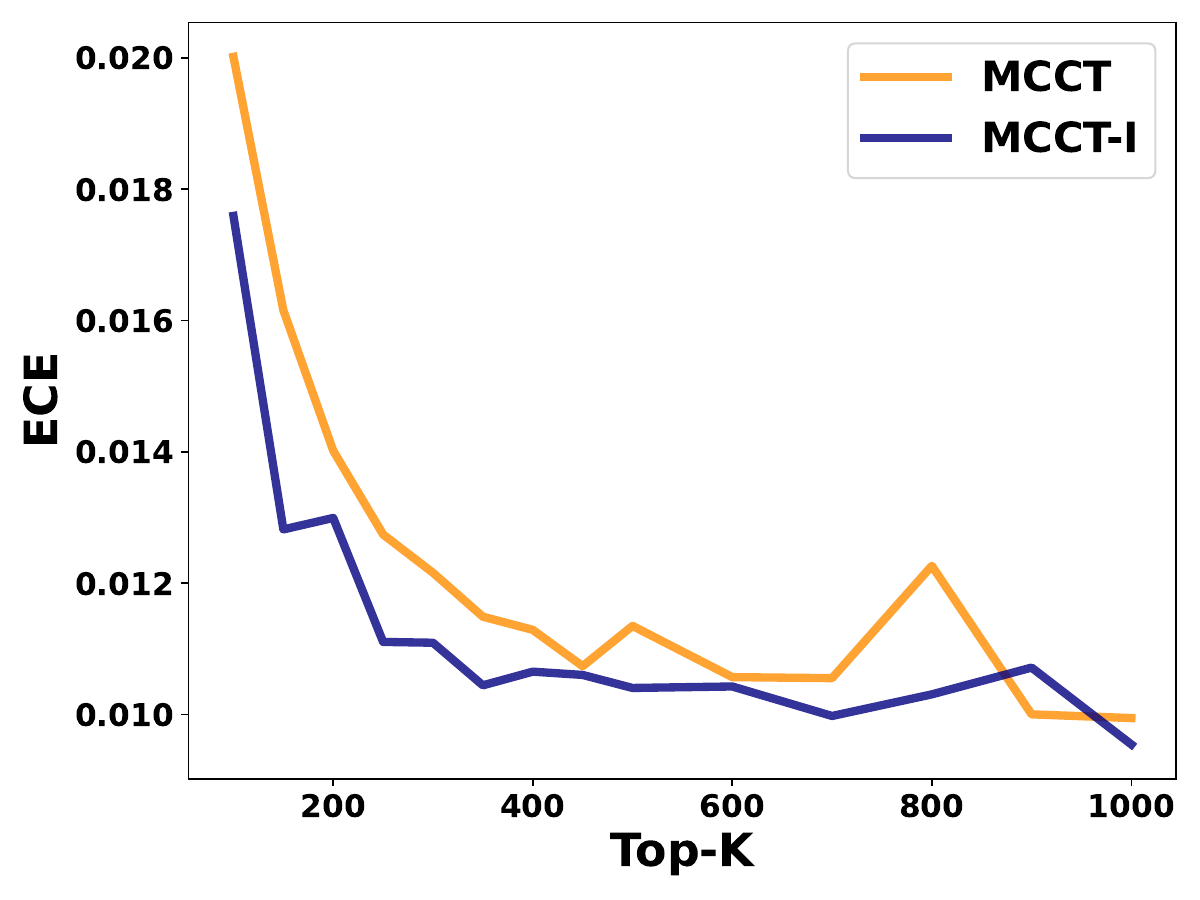}
    \label{fig:top_k_ECE_50}}
    \subfloat[ResNet152]{\includegraphics[width= .5\columnwidth]{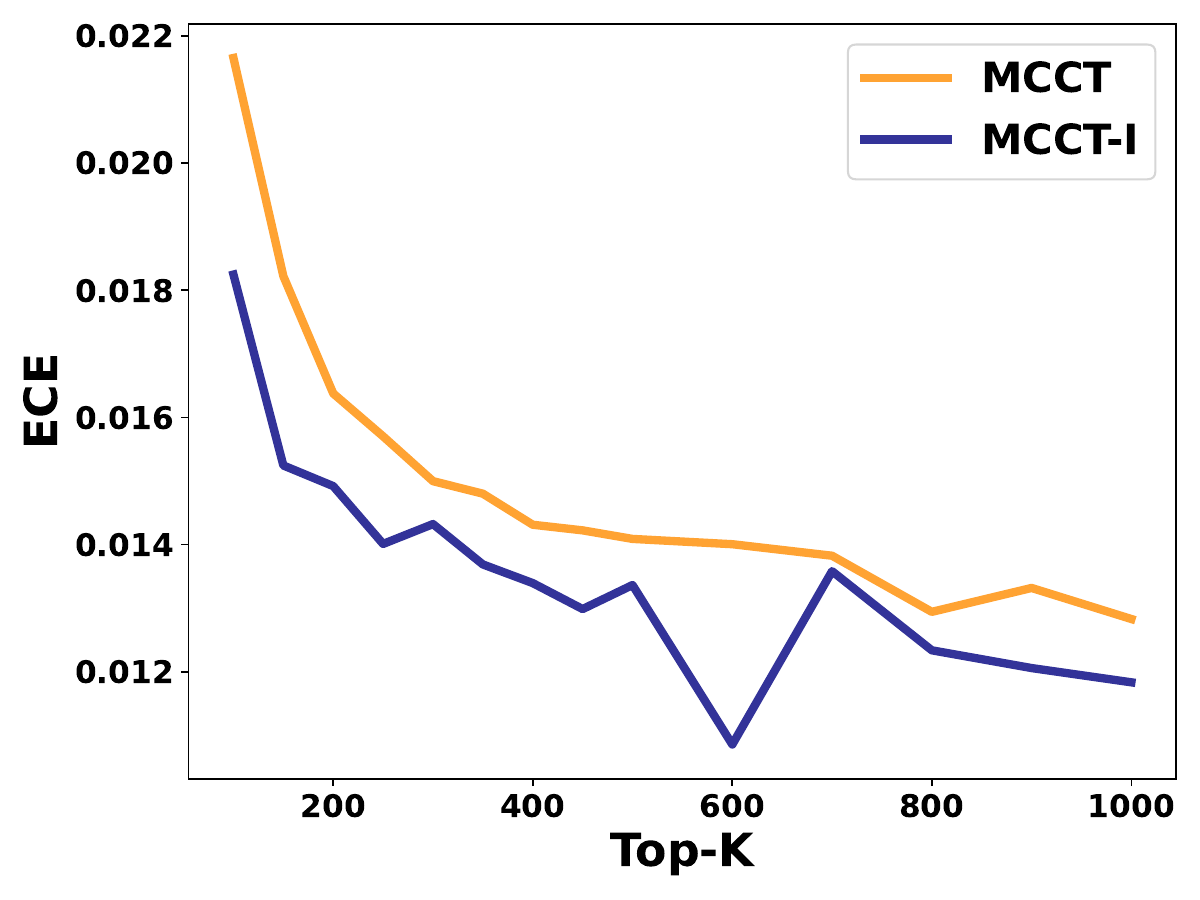}
    \label{fig:top_k_ECE_125}}
    \caption{The impact of top-$k$ logit selection on ECE. The x-axis represents the top-$k$ value, and the y-axis represents ECE. (a) ImageNet with ResNet-50, (b) ImageNet with ResNet-152.}
    \label{fig:top-k_ECE}
\end{figure}

Figure \ref{fig:top-k_ECE} presents the plot between ECE and the top-$k$ logit. For ImageNet-1K, both MCCT and MCCT-I maintain stable performance in terms of ECE until the top-$k$ parameter is reduced below 300. When the top-$k$ falls below 200, performance begins to deteriorate rapidly. Consistent with our previous results, MCCT-I generally outperforms MCCT in datasets with many classes. 

This result confirms our previous assumption that lower-ranked logits contain little helpful information for calibration. For datasets with a large number of classes, the probability distribution follows a probability simplex with a degree of freedom of $m-1$, and due to the NLL loss used in neural network models, classifiers tend to produce a long-tailed top-1 probability distribution. As a result, lower-ranking probability outputs usually have tiny values, which makes the corresponding logits contribute no information for calibration. Thus, our approach, which involves sorting logits and discarding the lower-ranking logits to a certain extent, does not significantly degrade calibration performance. 

\begin{figure}[htbp]
    \centering
    \subfloat[ResNet50]{\includegraphics[width= .5\columnwidth]{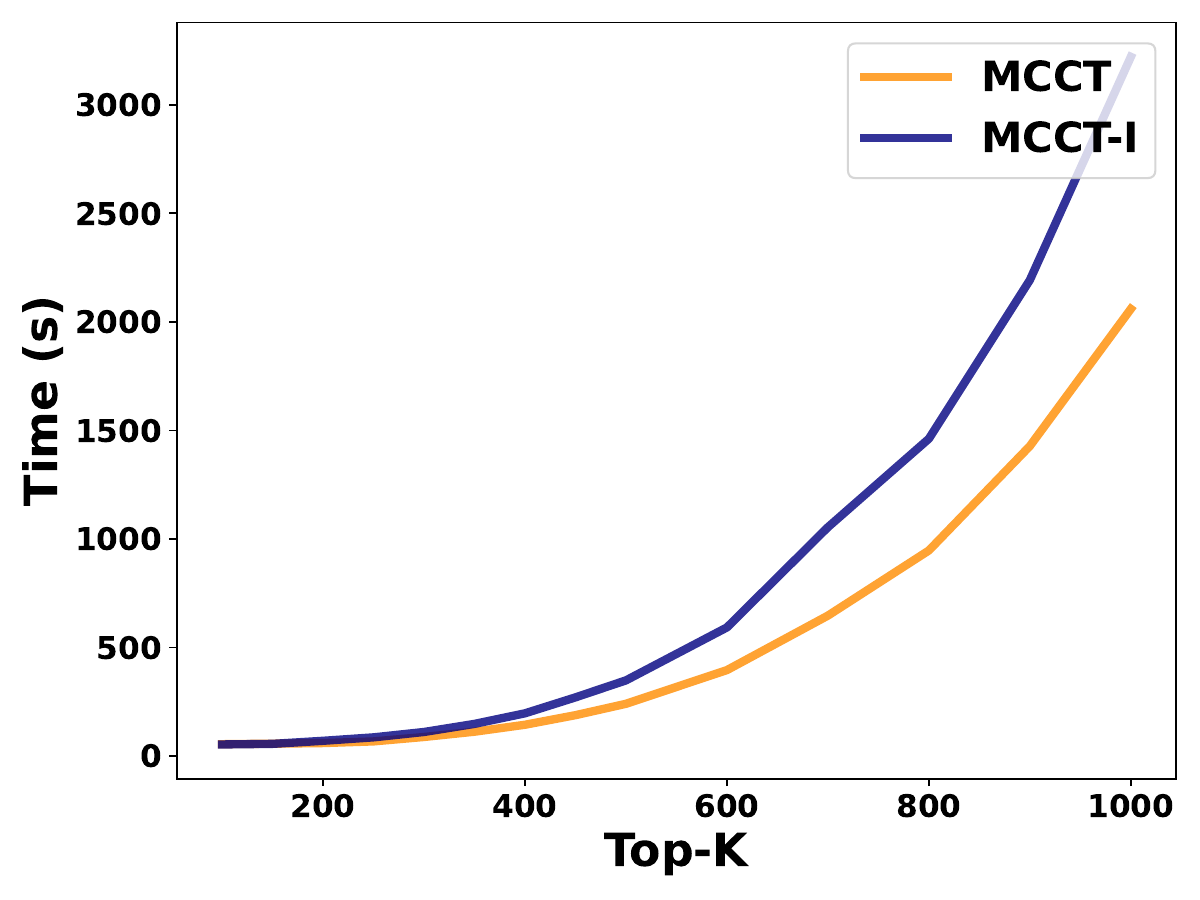}
    \label{fig:top_k_time_50}}
    \subfloat[ResNet152]{\includegraphics[width= .5\columnwidth]{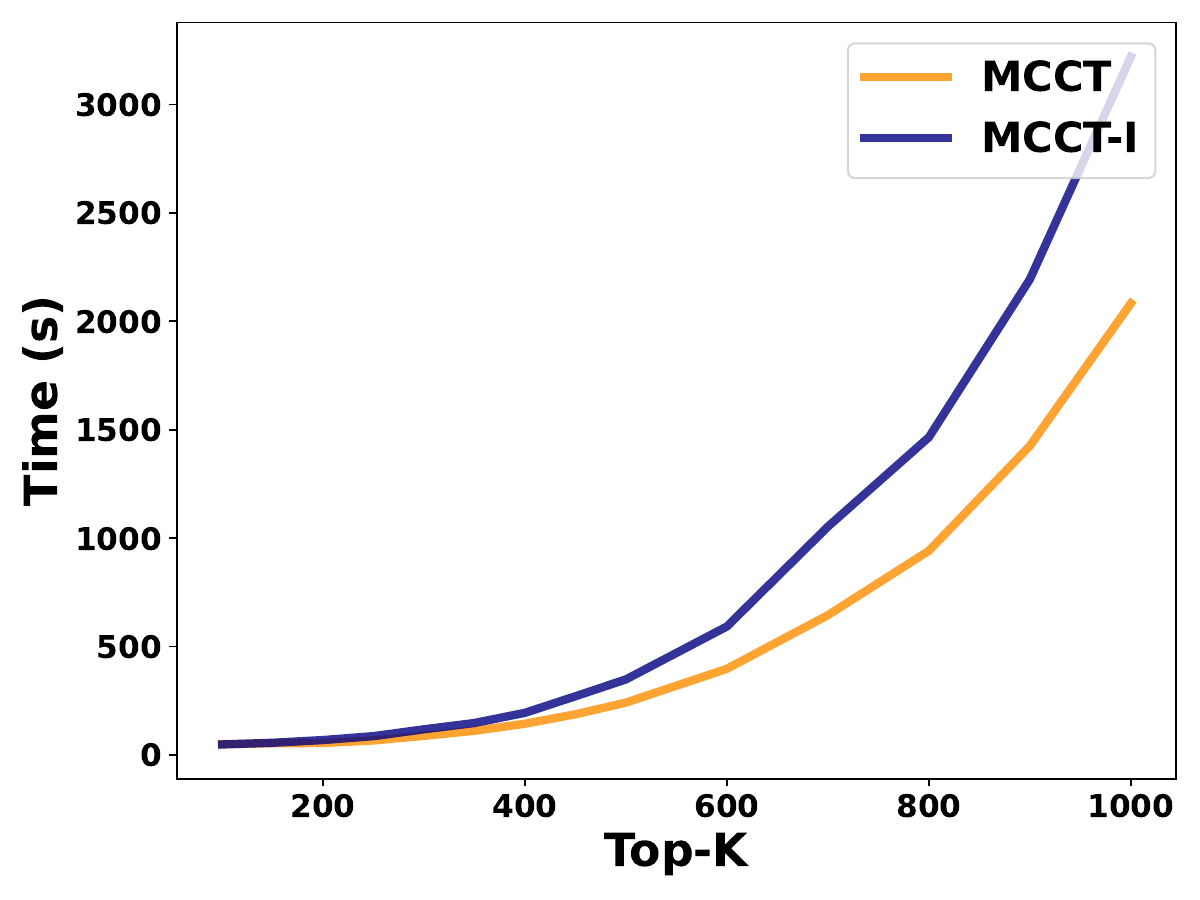}
    \label{fig:top_k_time_152}}
    \caption{The impact of top-$k$ logit selection on training time. The x-axis represents the top-$k$ value, and the y-axis represents training time in seconds. (a) ImageNet with ResNet-50, (b) ImageNet with ResNet-152.}
    \label{fig:top_k_time}
\end{figure}

Another interesting phenomenon we observed during the top-\(k\) experiment relates to the interpretability aspect of MCCT and MCCT-I. When examining the fitted parameters \(w\) and \(b\) of MCCT and MCCT-I across the full 1000 class ImageNet, we find that there exists a rank beyond which the parameters stabilize and cease to change. Notably, this stabilization point aligns closely with the elbow point observed in Figure \ref{fig:top-k_ECE}. This observation provides insight into when the sorted logits begin to resemble noise and no longer contribute meaningful information, thereby shedding light on model behavior in the many-class setting.

By reducing the number of top-$k$ logits, we observe a significant decrease in training time for both MCCT and MCCT-I. As shown in Figure \ref{fig:top_k_time}, the training time is reduced to under 10 minutes when top-$k$ is set to 600 and further drops to 3 minutes and 14 seconds when top-$k$ is reduced below 400. In practical applications, we recommend using all logits for training MCCT and MCCT-I on datasets with fewer than 500 classes. However, for datasets with more than 500 classes, a top-$k$ scheme that retains the top 400 logits can be used for both training and inference to improve efficiency without significantly affecting performance.

\section{Conclusion}

In this paper, we introduce a family of novel calibration methods for monotonic post-hoc calibration. By proposing a monotonic calibration map that is parameterized linearly with respect to the number of classes and has an intuitive interpretation of the parameters, the introduced calibration method is both expressive and interpretable. Furthermore, by formulating the implementation as a constrained optimization problem, the proposed method achieves state-of-the-art performance across various datasets and neural network models, outperforming previous post-hoc calibration methods such as ETS, PTS and DIAG while being data-efficient and robust in the sense of prediction and ranking preservation.

% \begin{contributions} % will be removed in pdf for initial submission 
% 					  % (without ‘accepted’ option in \documentclass)
%                       % so you can already fill it to test with the
%                       % ‘accepted’ class option
%     Briefly list author contributions. 
%     This is a nice way of making clear who did what and to give proper credit.
%     This section is optional.

%     H.~Q.~Bovik conceived the idea and wrote the paper.
%     Coauthor One created the code.
%     Coauthor Two created the figures.
% \end{contributions}

% \begin{acknowledgements} % will be removed in pdf for initial submission,
% 						 % (without ‘accepted’ option in \documentclass)
%                          % so you can already fill it to test with the
%                          % ‘accepted’ class option
%     Briefly acknowledge people and organizations here.

%     \emph{All} acknowledgements go in this section.
% \end{acknowledgements}

% References

\bibliography{uai2025-template}

\newpage

\onecolumn

% \title{Supplementary Material}
% \maketitle
\appendix

\section{Supplementary Material}

\setcounter{equation}{0}

\subsection{Gradient of MCCT and MCCT-I parameters}

\subsubsection{MCCT}

First, let's revisit the calibration map defined in Theorem \ref{the1} we have:
\begin{equation}
    f(Z) = S(Z) \odot W + b
    \label{eq:calmap1}
\end{equation}

For the purpose of calculating the gradient, we could ignore the constraints. To simplify the notation, instead of reverse sorting the transformed logits, here we remove the $S^{-1}$ the inverse sorting operation, and instead of calculating the loss between the S(Z) and the permuted one-hot encoded label $Y$ that each vector $Y$ is permuted in accordance to the sorting index of the corresponding logit vector $Z$.
 \begin{equation}
      \hat{p}_j = \frac{e^{f(Z)_j}}{\sum_{i=1}^{n} e^{f(Z)_i}}
 \end{equation}
 
The transformed logits are then go through softmax operation to produce the probability estimation $P$.
\begin{equation}
    L = -\sum_{j=1}^{n} Y_j \log  \hat{p}_j
    \label{loss}
\end{equation}
And the loss $L$ is calculated by taking the  Negative log-likelihood loss between the transform probability $P$ and the sorted one-hot encoded logit $Y$.

\begin{equation}
\frac{\partial L}{\partial f(Z)} =  \hat{p} - Y
\label{logit_loss}
\end{equation}

Equation \ref{logit_loss} shows the gradient vector of \( L \) w.r.t. \( f(Z) \), this is the well known derivative of the Softmax Function combine with the Categorical Cross-Entropy Loss.

\begin{equation}
    \frac{\partial f(Z)}{\partial W} = S(Z)
    \label{eq:zw}
\end{equation}

Equation \ref{eq:zw} describes the gradient of \( f(Z) \) w.r.t. \( W \).

\begin{equation}
\frac{\partial L}{\partial W} = \frac{\partial L}{\partial f(Z)_j} \cdot \frac{\partial f(Z)}{\partial W} =  S(Z) \odot (\hat{p}-Y)
\label{eq:lt}
\end{equation}

Then by using the chain rule we have the gradient vector of \( L \) w.r.t. \( W \) described in Equation \ref{eq:lt}

\begin{equation}
    \frac{\partial L}{\partial B} = \frac{\partial L}{\partial f(Z)} \cdot\frac{\partial f(Z)}{\partial B}
    = (\hat{p}-Y)
    \label{eq:bias1}
\end{equation}

From equation \ref{eq:calmap1} we know that the $\frac{\partial f(Z)}{\partial B}$ is simply $1$. Then by using the chain rule we have the gradient vector of \( L \) w.r.t. \( B \) described as in Equation \ref{eq:bias1}

\subsubsection{MCCT-I}

For the calibration map described in the Theorem \ref{the2} we have 

\begin{equation}
    f(Z) = \frac{S(Z)}{W} + b
\end{equation}

We can see that the $\frac{\partial L}{\partial B}$ is the same as the first calibration map.

\begin{equation}
    \frac{\partial f(Z)}{\partial W} = - \frac{S(Z)}{W^2}
    \label{eq:zw2}
\end{equation}

Equation \ref{eq:zw2} describes the gradient of \( f(Z) \) w.r.t. \( W \). 

\begin{equation}
\frac{\partial L}{\partial W} = \frac{\partial L}{\partial f(Z)} \cdot \frac{\partial f(Z)}{\partial W} = -\frac{S(Z)}{W^2} \odot (\hat{p}-Y),
\label{eq:lt2}
\end{equation}

Then by using the chain rule we have the gradient vector of \( L \) w.r.t. \( W \) described in Equation \ref{eq:lt2}

% \section{Additional Experiment Detail}
% In this section we provide additional details for our experiment.

\subsection{Additional information for Base Classifier}
For CIFAR-10 and CIFAR-100, we use the logits provided by previous work \citep{kull_beyond_2019}, available at \url{https://github.com/markus93/NN_calibration}. For ImageNet-1K, we use the PyTorch pre-trained models \citep{paszke2019pytorch} from \url{https://pytorch.org/vision/main/models.html} and perform inference on the original ImageNet-1K test set, which is randomly divided 50-50 into a calibration set and a testing set.

\end{document}